\documentclass[conference]{IEEEtran}
\IEEEoverridecommandlockouts
\usepackage{cite}
\usepackage{amsmath,amssymb,amsfonts}
\usepackage{algorithmic}
\usepackage{graphicx}
\usepackage{textcomp}

\def\BibTeX{{\rm B\kern-.05em{\sc i\kern-.025em b}\kern-.08em
    T\kern-.1667em\lower.7ex\hbox{E}\kern-.125emX}}
\begin{document}
\raggedbottom

\title{Weather Classification: A new multi-class dataset, data augmentation approach and comprehensive evaluations of Convolutional Neural Networks \\
\thanks{This work is supported by the UK Engineering and Physical Sciences Research Council through grants EP/R02572X/1 and EP/P017487/1}}
\author{\IEEEauthorblockN{Jose Carlos Villarreal Guerra\textsuperscript{1}, Zeba Khanam\textsuperscript{1}, Shoaib Ehsan\textsuperscript{1}, Rustam Stolkin\textsuperscript{2}, Klaus McDonald-Maier\textsuperscript{1}}
	\IEEEauthorblockA{\textsuperscript{1}\textit{Embedded and Intelligent Systems Lab, University of Essex} \\
		\textit{\textsuperscript{2}Extreme Robotics Lab (ERL), University of Birmingham}\\
		\{ \textsuperscript{1}zeba.khanam, \textsuperscript{1}sehsan, \textsuperscript{1}kdm \}@essex.ac.uk,
		\textsuperscript{1} josecarlosvillarrealguerra@gmail.com,
		\textsuperscript{2} r.stolkin@cs.bham.ac.uk
	}
}

\maketitle

\begin{abstract}
Weather conditions often disrupt the proper functioning of transportation systems. Present systems either deploy an array of sensors or use an in-vehicle camera to predict weather conditions. These solutions have resulted in incremental cost and limited scope. To ensure smooth operation of all transportation services in all-weather conditions, a reliable detection system is necessary to classify weather in wild. The challenges involved in solving this problem is that weather conditions are diverse in nature and there is an absence of discriminate features among various weather conditions. The existing works to solve this problem have been scene specific and have targeted classification of two categories of weather. In this paper, we have created a new open source dataset consisting of images depicting three classes of weather i.e rain, snow and fog called RFS Dataset. A novel algorithm has also been proposed which has used super pixel delimiting masks as a form of data augmentation, leading to reasonable results with respect to ten Convolutional Neural Network architectures.
\end{abstract}

\begin{IEEEkeywords}
Weather Classification , Convolutional Neural Network , Superpixels, Data Augmentation 
\end{IEEEkeywords}

\section{Introduction}
\label{sec:intro}

Time and again unfortunate accidents due to inclement weather conditions across the globe have surfaced. Ship collision, train derailment, plane crash and car accidents are some of the tragic incidents that have been a part of the headlines in recent times. This grave problem of safety and security in adverse conditions has drawn the attention of the society and numerous studies have been done in past to expose the vulnerability of functioning of transportation services due to weather conditions \cite{andrey2003weather}. In past, weather-controlled driving speeds and behaviour have been proposed \cite{rama2004effects}. With the advancement in technology and emergence of a new field, intelligent transportation, automated determination of weather condition has become more relevant. Present systems either rely on series of expensive sensors or human assistance to identify the weather conditions \cite{christensen1966investigation}, \cite{nystuen1997weather}, \cite{schnur1998case}. Researchers, in recent era have looked at various economical solutions. They have channelled their research in a direction where computer vision techniques have been used to classify the weather condition using a single image. The task of assessing the weather condition from a single image is a straightforward and easy task for humans. Nevertheless, this task has a higher difficulty level for an autonomous system and designing a decent classifier of weather that receives single images as an input would represent an important achievement.

The work described in this paper translates to two contributions to the field of weather classification. The first one is exploring the use of superpixel masks as a data augmentation technique, considering different Convolutional Neural Network (CNN) architectures for the feature extraction process when classifying outdoor scenes in a multi-class setting using general-purpose images. The second contribution is the creation of a new, open source dataset, consisting of images collected online that depict scenes of three weather conditions, called Rain Fog Snow \textbf{(RFS)} dataset.

The paper has been organized into six sections. In the section \ref{sec:intro}, a brief introduction to the topic has been presented. This has been followed by the section \ref{sec:rw} which narrates the recent advances made in this field. Section \ref{sec:data} presents detail description of proposed dataset \textit{'RFS'}. Section \ref{meth} mentions the corresponding background on the use of superpixel masks and details of chosen architectures for evaluation.  Section \ref{meth}  also reports description of the methodology, emphasizing on the employed pipeline and other parameters that were used. Section \ref{res} evaluates the results of experiment and inferences that can be drawn. Finally, section \ref{con} sheds a light on conclusions obtained that can guide future work that is relevant to the topic.

\section{Related Work}
\label{sec:rw}
In recent years, important contributions have been made as an attempt to solve the weather classification problem. Many of these recognisable attempts target the problem from perspective of weather classification for traffic purposes were limited to single adverse condition like rain(\cite{roser2008classification}, \cite{kurihata2005rainy}, \cite{cord2014detecting},\cite{kuehnle2013method}) and fog(\cite{pavlic2013classification}, \cite{belaroussi2015convergence}, \cite{pavlic2016method}). The dataset used to train the classifier in this problem contains images captured using on board camera of various weather conditions on roads and highways. This implies that they share a set of features that are exclusive to specific purpose of driving assistance on road and cannot not be generalised for the classification of images with different backgrounds and viewpoint. 

The generalization of this task has added to complexity of the problem as it is hard to find a consistent set of features that can be extracted in all the relevant images. Differing contributions have been made on the most relevant features for weather classification. Lu et al. \cite{lu2014two} proposed a two-class weather classifier which classified images based on five features (Sky, Haze, Contrast, Reflection and Shadow). This work was extended by the authors by contacting CNN features in the feature vector. Similarly, Zhang and Ma \cite{zhang2015multi} rely on manual feature identification and extraction for the development of their weather classifier. They use global and local features to represent the images. The authors refer to local features, which are the set of characteristics that are used to represent a given type of weather, and thus would not be present in images depicting a different kind of weather. For a more generalised visualisation, they use global features which are presented as general characteristics of images that could be present in images with any kind of weather. 

Convolutional Neural Networks (CNNs) have been used in recent works for image classification. Krizhevsky et al. \cite{krizhevsky2012imagenet} implemented the special architecture of CNNs for image classification, taking advantage of the advances in computing power needed for its training and overall performance. Their implementation showed the capabilities of an image classifier with a built-in feature extraction based on supervised learning \cite{krizhevsky2012imagenet}, instead of the manual extraction of features used in other machine learning applications.

In the field of weather classification, CNNs have been used in \cite{elhoseiny2015weather} and \cite{zhu2016extreme}. In their paper, Elhoseiny et al. \cite{elhoseiny2015weather} created a network based on the work by Krizhevsky et al. \cite{krizhevsky2012imagenet} for the categorization of images between two possible classes. Their contribution represented an insightful approach that adopted the technology being used for general purposes into a highly specific application, proving that it can achieve exceptional results. Lu et al. \cite{7784804} extended their work of two-class weather recognition by adding CNN features used by \cite{elhoseiny2015weather}. Similarly, Zhu et al. \cite{zhu2016extreme} explored an implementation that has proven successful for the task of image classification, which is the architecture known as GoogLeNet, proposed in \cite{szegedy2015going}. This implementation of CNNs had more layers than its counterparts at the moment of its proposal and achieved better results in the experimentation carried out in \cite{zhu2016extreme} used on a collection of weather images generated by the same team. Di Lin et al. \cite{lin2017rscm} proposed a deep learning framework named region selection and concurrency model (RSCM) which used regional cues to predict the weather condition.

\section{RFS Dataset}
\label{sec:data}

\begin{figure}

\begin{minipage}[b]{1.0\linewidth}
  \centering
  \centerline{\includegraphics[width = \linewidth]{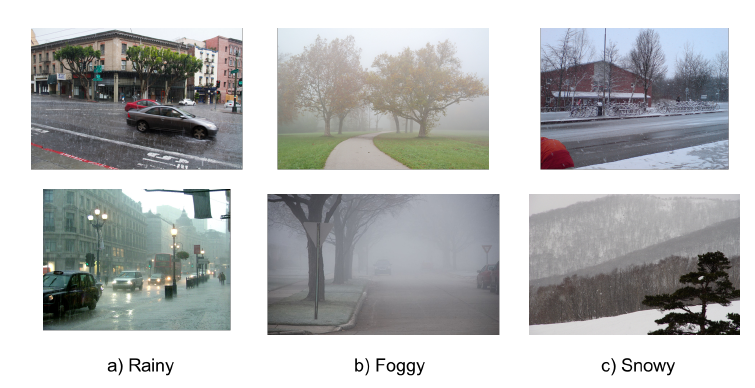}}
\end{minipage}

\caption{Sample images of all three categories of RFS Weather Dataset. }
\label{fig:1}
\end{figure}

For the purposes of creating an open source dataset that can potentially contribute to future efforts in the field of computer vision, the images of interest containing Creative Commons \cite{WhenWeShare} license retrieved from Flickr \cite{flickr}, Pixabay\cite{free} and Wikimedia Commons\cite{ wiki:xxx} have been used. Different images are subject to different types of licenses; however, a common requirement is to give attribution to the author. This premise as a method to collect images is a fair way to build a dataset intended to be shared while respecting the producers of the content of interest. Unfortunately, the amount of images with the Creative Commons licenses are considerably smaller than the number of images with reserved copyright, leading to potentially smaller datasets. The collection of images created as part of this work composes the RFS Weather Dataset\footnote{RFS can be accessed online using the link: https://github.com/ZebaKhanam91/SP-Weather} , named after the three categories that it includes. Figure \ref{fig:1} shows a sample of the images contained in RFS Weather Dataset.

\subsection{Images of Fog}
\label{fog}

The task to look for images depicting foggy weather was
relatively easy, as the different image hosting web sites have
a decent collection of this type of pictures. However, the
use of synonyms, as well as terms in other languages was
required to find the total number images that compose the
foggy category. Another set of useful search terms consists
on looking for locations (cities or towns) that are characterised
by foggy weather, and use these as the tags to look
for. The foggy category of the RFS Weather Dataset contains
1100 images.

\subsection{Images of Rain}
\label{rain}

Collecting images of rainy weather condition was a challenge. 
Since many of the images which were tagged as ``rainy'' on the 
platforms were representation of raindrops on glass scene and 
indoor activities for rainy days. However, 1100 quality images with 
landscapes and urban scenes of rainy days are represented in this category 
of the dataset. To gather the images, the first strategy is to use the word ”rainy” 
as the search term, collect the images and then move on to related words like ”rain” 
or ”deluge”, and finally use these terms in other languages.

\subsection{Images of Snow}
\label{snow}

The category for images showing snowy outdoor scenes
was built using terms in English, French and Spanish. The
terms included ”snowy” and ”snow”, and returned a decent
amount of images that are included in the dataset. From the
returned images, the 1100 most representative images are
included in the RFS Weather Dataset.

\section{Methodology}
\label{meth}

Cognitive psychology has been inspiration of many computer vision algorithms. In order to
contribute towards the development of the field, and particularly the weather classification task, an effort is presented
as an inquiry into the effect that superpixel delimiting masks on images have on the performance of weather classification.
Based on the recent success of Convolutional Neural Networks in the field of Computer Vision, different conceptions of this technology are used in the work described here. The purpose is to determine if any of these architectures can benefit from the use of superpixel masks.

The categories of interest for this contribution are  sunny, cloudy, foggy, rainy and snowy images. For the sunny and cloudy categories, Lu et al. \cite{lu2014two} developed a dataset as a part of their work. This dataset contains images from other datasets and images available online and labelled as part of their contribution. This dataset was made available on the web page associated with their paper. The images for the other categories are those belonging to RFS weather dataset. Since RFS weather Dataset contains 1100 images for each of its categories, 1100 images were taken from the categories in Lu et al. dataset, having a total of 5500 images distributed in five categories. To create a partition of the images for the Support Vector Machine classifier training and testing process, 70\% of the images are used for the training phase, while 30\% is used for the testing phase. However, the classifier is trained over each of the five different categories, which means that each class had a negative counterpart with images that served as an example of what the category does not look like (composed of images from the other classes of the combined datasets). With this dataset setting, the binary classifier for each category determine if a given image represents or not the current class.

\subsection{Pipeline}
\label{pipe}
\begin{figure}[h]

\begin{minipage}[b]{1.0\linewidth}
  \centering
  \centerline{\includegraphics[width = \linewidth, height = 4.5cm]{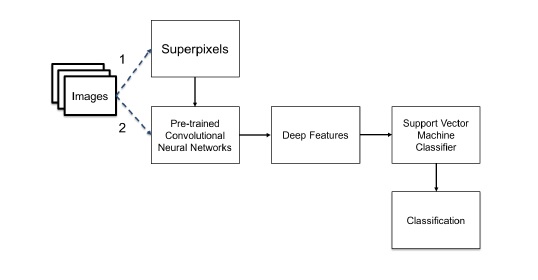}}
\end{minipage}

\caption{Project Pipeline}
\label{fig:3}
\end{figure}

\begin{figure}

\begin{minipage}[b]{1.0\linewidth}
  \centering
  \centerline{\includegraphics[width = \linewidth]{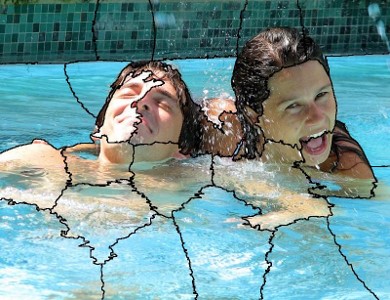}}
\end{minipage}

\caption{A sunny image with overlayed boundaries of 25 superpixel mask   }
\label{fig:4}
\end{figure}

The pipeline used for the work described in this paper is
based on the work by Kalliatakis et al. in [10]. A visual representation
of the pipeline is shown in Figure \ref{fig:3}. There are
two possible paths in the pipeline, either using superpixel
masks or using the raw images directly. In the first mode of
operation, the images are passed through a module that calculates
the superpixels, generates the mask in a colour
and applies it over the image, just as shown in Figure \ref{fig:4}. The
images, enhanced with supeprixel mask if used, are divided
into four partitions corresponding to positive training, positive testing, negative training, and negative testing. Then,
the matdeeprep function [10], which uses pre-trained Convolutional
Neural Network models, is used to extract deep
features from the images in all the partitions. Finally, the
features are used to train and test a binary Support Vector
Machine classifier.

\subsection{Software Implementation}
\label{soft}

The pipeline is implemented in MATLAB, coupled with
Caffe \cite{jia2014caffe}. The Matdeeprep function \cite{kalliatakis2017detection} is executed on top
of Caffe for the use of pre-trained models for the extraction
of features that are intended to be used in a classifier.
For the Support Vector Machine part of the pipeline, a special
library of Computer Vision algorithms was used, called
VLFeat and presented in \cite{vedaldi2012vlfeat}.

\subsection{Experimentation}
\label{exp}
The classifier was trained and tested on images of each category, which are cloudy, foggy,
rainy, snowy, and sunny. For each of the categories, 770
were used as positive training and testing set each, and 330 images were used for negative training and testing each.
After performing the different
runs of the experiment, the mean Average Precision
(mAP) was calculated for each of the models trained with
the results for each of the categories. The experimentation
proceeded in five different settings: using the raw images,
images with 25,50,75 and 100 superpixel masks.

\section{Results}
\label{res}

\begin{table}
\caption{Results from CNN evaluations} 
\begin{tabular}{|c | c | c | c | c | c |} 
\hline 
Model & 0 SP & 25 SP & 50 SP & 75SP & 100 SP \\
\hline 
CaffeNet &  0.7591 & 0.7568 & 0.7542 & 0.7564 & 0.7573  \\ 
\hline
PlacesCNN & 0.7627 & 0.7401 & 0.7311	& 0.7322 & 0.7335 \\
\hline
ResNet 50 & \textbf{0.7767}  &	\textbf{0.7851}  &	\textbf{0.7957} & \textbf{0.8007} & \textbf{0.8070} \\
\hline
ResNet 101 & 0.7681 &	0.7643 & 	0.7614 & 0.7698 & 0.7756 \\
\hline
ResNet 152 & 0.7504 & 0.7584  &	0.7538 &	0.7549 & 0.7519 \\
\hline 
VGG\_CNN\_F &  0.7134 & 0.7224 & 0.7252 &	0.7309	 & 0.7201 \\
\hline
VGG\_CNN\_M & 0.6849 & 0.6900 & 0.6947 &	0.6984 &	0.7006 \\
\hline
VGG\_CNN\_S & 0.7060 &	0.7202 &  0.7188  & 0.7229  & 0.7272 \\
\hline
VGGNet16 &  0.7194 & 0.7203 &	0.7252	& 0.7244 & 0.7218 \\
\hline
VGGNet19 & 0.6948 & 0.7008 &	0.7041 & 0.7111 & 0.7167 \\
\hline
\end{tabular}
\label{tab1} 
\end{table}

After the trial runs, the results show that the overall mAP for all of the models and settings is between 68\% and 81\%. Table \ref{tab1} shows the complete results of the experimentation process. The overall best classification was the ResNet 50 architecture for all of the settings of the superpixel masks. The three variations of the residual networks (ResNet) used in the experimentation were among the four top performing models in the four different settings. These results are a
good indication that the optimisation achieved by the inclusion of residual methods learning with shortcut connections \cite{he2016deep} has a positive effect in the overall task of weather classification, and it can benefit slightly from the use of superpixel masks as data augmentation. The results of this contribution are consistent with \cite{he2016deep}, paper in which Residual Networks are described as the winners of the 2015 ILSVRC competition, having a smaller error than GoogLeNet and the VGG networks. Residual Networks are thus not only successful in arbitrary image recognition, but can also work well as a solution to weather classification problem. Another notable fact that is evident with the results shown in Table \ref{tab1} is increase in performance between the experimental setting involving raw images and the ones involving superpixel masks for all models except for PlacesCNN model and CaffeNet model. 
\begin{figure}
\begin{minipage}[b]{1.0\linewidth}
  \centering
  \centerline{\includegraphics[width = \linewidth]{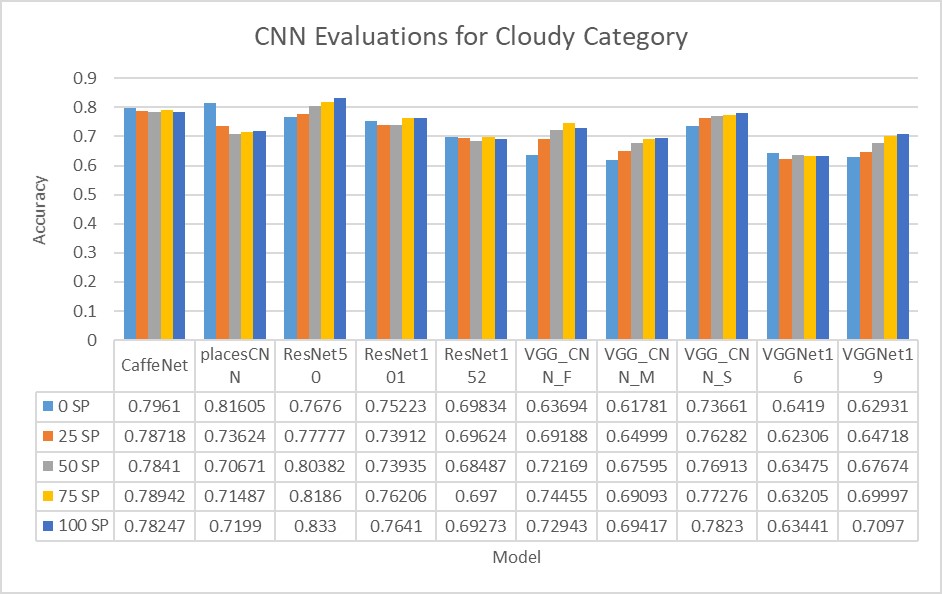}}
\end{minipage}

\caption{CNN Evaluations for Cloudy Category}
\label{fig:5}
\end{figure}
\begin{figure}
\begin{minipage}[b]{1.0\linewidth}
  \centering
  \centerline{\includegraphics[width = \linewidth]{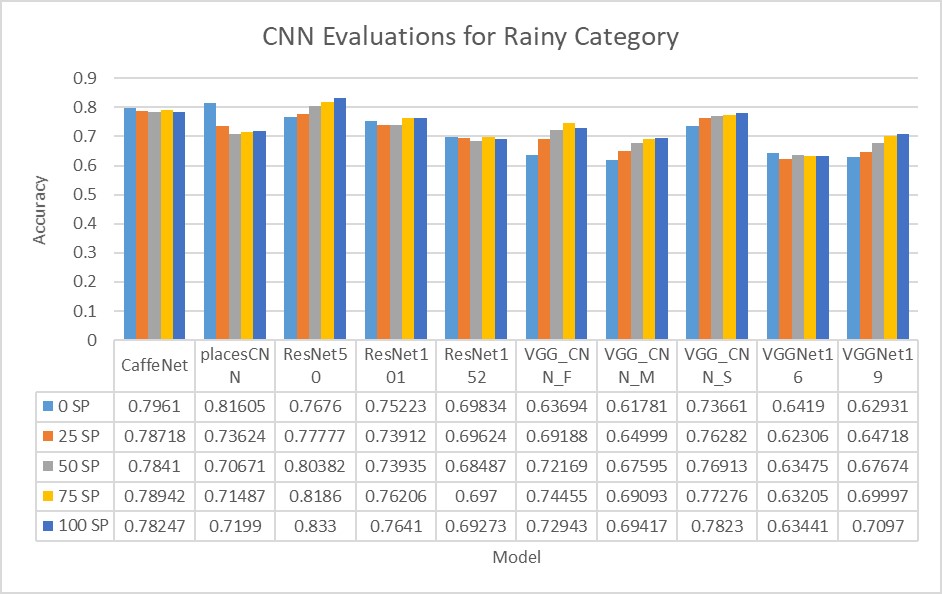}}
\end{minipage}

\caption{CNN Evaluations for Rainy Category}
\label{fig:6}
\end{figure}
\begin{figure}
\begin{minipage}[b]{1.0\linewidth}
  \centering
  \centerline{\includegraphics[width = \linewidth]{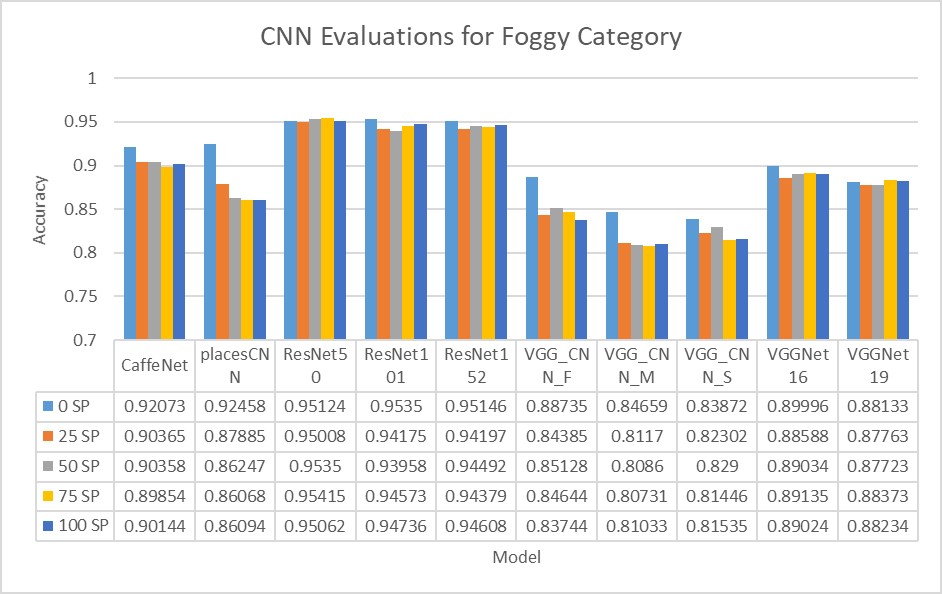}}
\end{minipage}

\caption{CNN Evaluations for Foggy Category}
\label{fig:7}
\end{figure}

\begin{figure}
\begin{minipage}[b]{1.0\linewidth}
  \centering
  \centerline{\includegraphics[width = \linewidth]{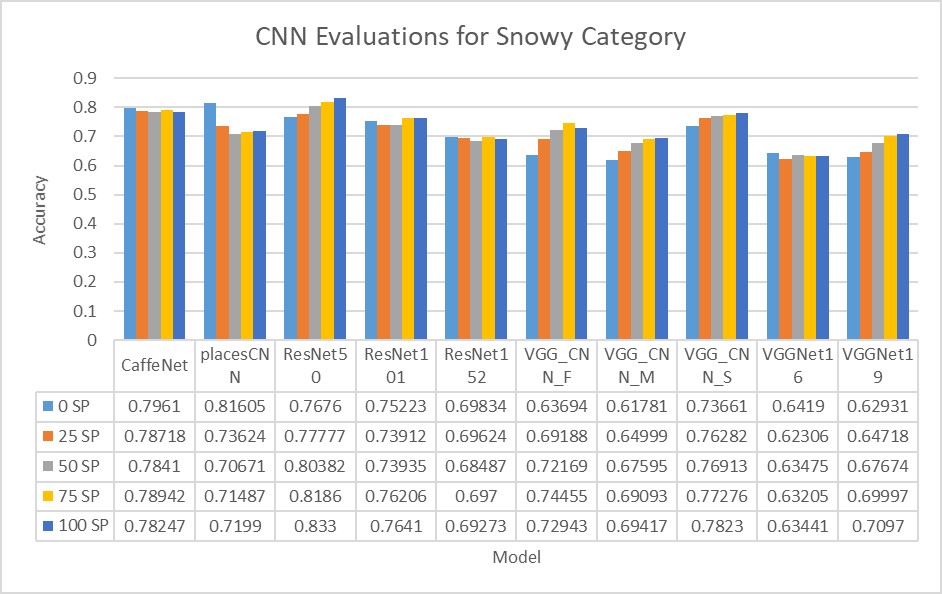}}
\end{minipage}

\caption{CNN Evaluations for Snowy Category}
\label{fig:8}
\end{figure}
\begin{figure}
\begin{minipage}[b]{1.0\linewidth}
  \centering
  \centerline{\includegraphics[width = \linewidth]{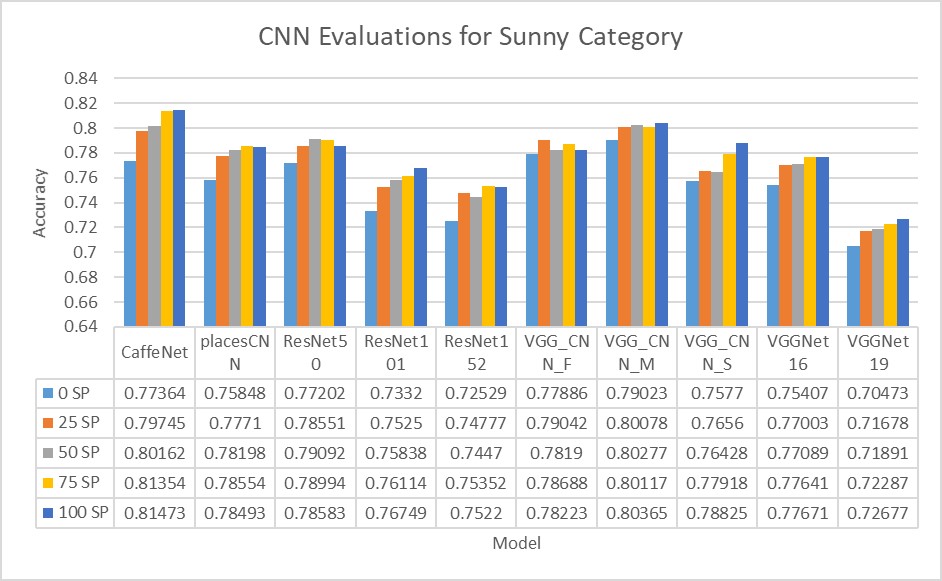}}
\end{minipage}

\caption{CNN Evaluations for Sunny Category}
\label{fig:9}
\end{figure}

The analysis of results category wise (Figure \ref{fig:5}-\ref{fig:9}) indicate that the best architecture to handle the weather classification problem are three ResNet architectures except for sunny category where VGG\_CNN\_M performs best. As described in \cite{he2016deep}, it is possible to use this type of networks with different depths. Concluding the potential of this type of network, draws attention for the possibility of more experimentation using them with different and larger datasets. Although certain improvement is noticeable, it is evident that computational cost will  be incurred on calculation of superpixel masks. The trade off between increase precision and computational cost is worth pursuing in future.  

\section{Conclusion}
\label{con}
This paper explores the possibility of using superpixel masks as a form of data augmentation that improves the performance of a classifier in the context of weather classification.  Also, a fair method for building datasets from online content is showcased, giving credit that corresponds to the creators of content. Proposed dataset is also hybrid in nature as images for two weather classes have been taken from a benchmark dataset. This allows persistent problem of biases in computer vision datasets to be tackled.   \\
The specific field of computer vision known as weather classification is still evolving and worthy proposals are still
being made to improve the baseline. An opportunity to advance the field even further lies in the adoption of new
techniques and tools that have been used to improve other fields. To achieve this, it is necessary to recognise that weather is a phenomenon that is a complex phenomena where semantic labelling of images can be a challenging task . A single image can encompass multiple weathers for instance partly cloudy. To exploit all the information that can be extracted from a single outdoor image, considering uncertainty as an opportunity rather than a problem is an important step. As Rutkowski states in \cite{rutkowski2006flexible}, different techniques of computer science sometimes can be combined to constitute new and better proposals, which might lead to implementations like neuro-fuzzy systems to tackle the weather classification problem.

\bibliographystyle{IEEEbib}
\bibliography{strings}

\end{document}